\documentclass[runningheads]{llncs}

\usepackage{eccv}

\usepackage{eccvabbrv}

\usepackage{graphicx}
\usepackage{booktabs}
\usepackage{multirow}
\usepackage{pifont}

\usepackage{amsthm,amsmath,amssymb}
\usepackage{mathrsfs}
\usepackage{wrapfig}

\usepackage[accsupp]{axessibility}  % Improves PDF readability for those with disabilities.

\usepackage[pagebackref,breaklinks,colorlinks,citecolor=eccvblue]{hyperref}
\usepackage{orcidlink}
\newcommand{\myPara}[1]{\vspace{.05in}\noindent\textbf{#1}\quad}

\begin{document}

\title{Self-Explainable Affordance Learning with Embodied Caption} 

\titlerunning{Self-Explainable Affordance Learning}

\author{Zhipeng Zhang\inst{1,2} \and
Zhimin Wei\inst{1} \and \\
Guolei Sun\inst{2}\and Peng Wang\inst{1}\thanks{*Corresponding author.} \and Luc Van Gool\inst{2}}

\authorrunning{Zhang.~Author et al.}

\institute{School of Computer Science, Northwestern Polytechnical University\and
Computer Vision Lab, ETH Zurich
}

\maketitle

\begin{abstract}
  In the field of visual affordance learning, previous methods mainly used abundant images or videos that delineate human behavior patterns to identify action possibility regions for object manipulation, with a variety of applications in robotic tasks. However, they encounter a main challenge of action ambiguity, illustrated by the vagueness like whether to beat or carry a drum, and the complexities involved in processing intricate scenes. Moreover, it is important for human intervention to rectify robot errors in time. To address these issues, we introduce Self-Explainable Affordance learning (SEA) with embodied caption. This innovation enables robots to articulate their intentions and bridge the gap between explainable vision-language caption and visual affordance learning. Due to a lack of appropriate dataset, we unveil a pioneering dataset and metrics tailored for this task, which integrates images, heatmaps, and embodied captions. Furthermore, we propose a novel model to effectively combine affordance grounding with self-explanation in a simple but efficient manner. Extensive quantitative and qualitative experiments demonstrate our method's effectiveness.
\end{abstract}

\section{Introduction}
\label{sec:intro}

The advent of embodied intelligence~\cite{driess2023palm,vemprala2023chatgpt,brohan2023rt,padalkar2023open,jin2023alphablock,lynch2023interactive,mu2023embodiedgpt} has marked a significant research juncture, synergizing computer vision, natural language processing, and robotic control. The concept of visual affordance learning~\cite{ardon2020affordances,bahl2023affordances,liu2022joint,nagarajan2019grounded,goyal2022human,li2023locate,chuang2018learning,mees2023grounding} has emerged as a key strategy to merge visual perception with robotic actuation. 
This paradigm facilitates the robotic acquisition of actionable insights without the prerequisite of voluminous offline reinforcement learning or a large number of robot-specific annotated data. Such advancements permit robots to assimilate and execute actions by merely observing human-object interactions. It also mitigates the challenges associated with collecting data for instruction-conditional robotic datasets. 

\begin{figure}[t]
  \centering
   \includegraphics[width=1\textwidth]{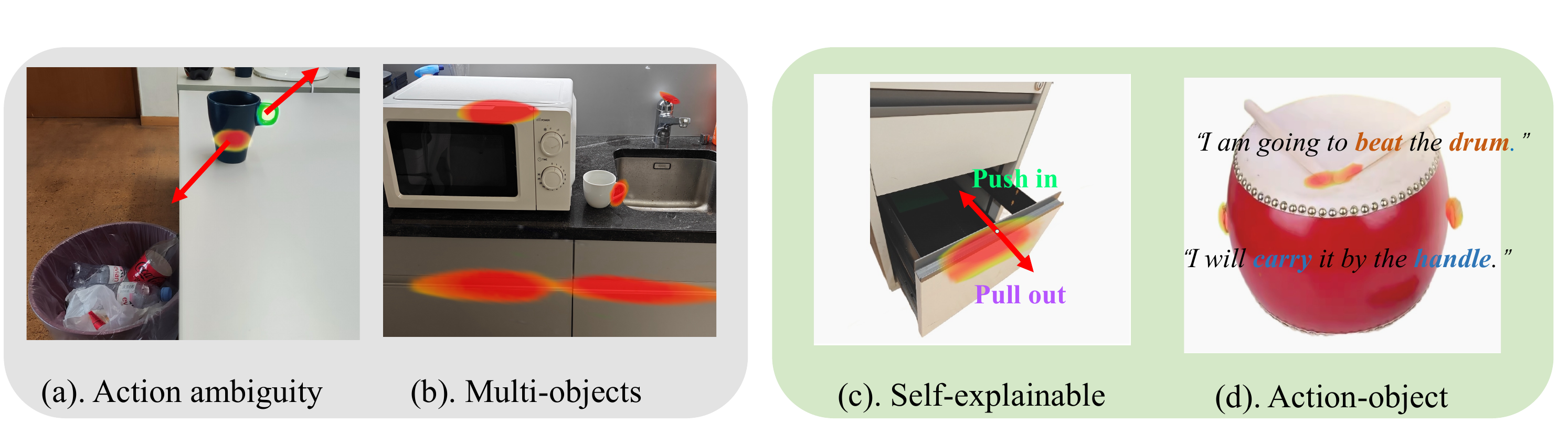}
   \vspace{-0.3cm}
   \caption{ Existing visual affordance learning suffers from the challenges (a-b) and solutions (c-d). (a) \textit{Action ambiguity}: Pick it up to drink? Pour it down? %Different goal actions correspond to different perceptual heat points and trajectories. 
   (b) \textit{Multi-objects ambiguity}: Each heat point corresponds to a distinct behavioral purpose. To tackle these issues, we introduce self-explainable affordance learning. In (c), the robot can say which action it is going to do. In (d), objects intended for interaction are distinguished with clear and defined descriptions.}
   \label{fig1}
   \vspace{-0.3cm}
\end{figure}

However, grounding affordance regions directly from egocentric visual input poses significant challenges. Previous research\cite{li2023locate,luo2023grounded,liu2022joint} often relied on learning exocentric image interaction features to predict heat points or action trajectories, such as grasping, pushing, and pulling. Visual affordance learning mainly employs large-scale exocentric visual data involving human-object interactions images or videos. However, there are ambiguity problems in previous visual affordance learning. As shown in Fig.~\ref{fig1}, action ambiguity and multi-region ambiguity problems exist. Specifically, if there is a cup of water on the table, should the agent lift the cup handle upwards to drink or tilt its body downwards to pour? In the case of a drum, should the agent beat the drumhead or carry it by the handle? These cases may appear in exocentric training data. Thus there is a possibility that all of these predictions are correct when confronted with egocentric pictures. These challenges highlight the intricate interconnections between objects, actions, and affordance maps. A significant gap exists due to a lack of datasets with corresponding captions that can address these ambiguities and provide explanatory descriptions for the robotic operations. 

To address this challenge,
we introduce a novel paradigm of self-explainable affordance learning with embodied captions, motivated by the necessity for robots to perceive and express their intended objects/actions in a straightforward way that humans can understand. We require that the model not only predict affordance maps from a visual perspective but also produce corresponding object-action descriptions as embodied caption. This joint vision-language (VL) affordance learning will generate more accurate and interpretable affordance results to reduce ambiguity. 

To support this novel task, we have developed a self-explainable affordance learning dataset with manually annotated descriptions of actions and objects. Based on the robotic visual affordance dataset from AGD20K~\cite{luo2022learning}, we conducted rigorous filtering of unfavorable images and then annotated high-quality object-action descriptions (captions). Specifically, each image underwent a rigorous review process by multiple annotators to ensure clarity and resolve any action ambiguities. Then annotation experts labelled the exact object-action descriptions corresponding to the interaction region of each image. Next, annotators performed cross-checks to verify the accuracy of critical information, including location, object, and action. 
Therefore, it facilitates the model to excel both the low-level task of visual feature localization and the high-level tasks of vision-language semantics correlation. These are not disjoint subtasks but intertwined components of a cohesive system. 
Within the realm of self-explainable embodied captions for robotics, actions serve as `attribute labels', diverging from traditional object grounding. For example, a drum may present different action regions for `beating' and `carrying' in Fig.~\ref{fig1} (d). 
% Simply merging these two aspects without nuanced integration could lead to confusion due to the multiplicity of potential interactions. 

To effectively address these challenges, we propose the Self-Explainable Affordance model, which innovatively integrates affordance prediction and caption generation. Unlike simple baseline attempts that treat these as directly compatible subtasks, we adopt a more complex and synergistic approach. It generates object-action descriptions that are deeply rooted in the understanding of the visual scene context.
It then enhances VL affordance grounding accuracy by leveraging text features associated with the object-action pair. Additionally, our alignment branch loss computation is specially designed to match the localization region's features with the corresponding text features. This method allows our model to effectively combine and complement the aspects of visual scene understanding and targeted action description.

To evaluate self-explainable affordance learning with embodied caption across a range of heterogeneous agents, we have developed a novel metric: self-explainable embodied caption accuracy. The realm of intelligent agents is vast and varied~\cite{yang2022motivating,ishiguro2001robovie,kaneko2019humanoid,yamano2005five,chao2019deep,handa2020dexpilot,fu2020d4rl,brohan2022rt,brohan2023rt,kornatowski2020morphing,meem2022semi}, each contributing to a diverse array of syntactic prompt/descriptions. Despite this diversity of expression style, the crucial element is the object-action relationship. Therefore, we introduce the top-k object-action accuracy metric, designed to evaluate vital object-action information from the robot's self-explainable embodied descriptions. %Specifically, top-1 result corresponds the maximum probability visual heat map prediction. 
For evaluation of affordance grounding, the heat map associated with highest probability of object-action prediction is used.
%This metric significantly bolsters the generalization capacity of our dataset and tasks, facilitating the open evaluation of various models and agents, even in scenarios akin to open vocabulary classification.
%This metric effectively evaluate our self-explainbale affordance learning, even in scenarios akin to open vocabulary classification.
This metric provides an excellent evaluation criterion for models designed for self-explainable affordance learning task.

In conclusion, our study makes the following main contributions:\\
(1) We introduce the innovative concept of self-explainable affordance learning with embodied caption. This novel task challenges the model not only to learn touchable heat points from extensive exo images to ego images, but also to articulate the actions and objects involved.\\
(2) To facilitate this task, we have developed a high-quality self-explainable affordance learning dataset. 
This dataset enables agents to perform affordance grounding and offer corresponding explainable descriptions. Additionally, we have developed a new key behavior metric tailored for assessing self-explainable visual affordance learning.\\
(3) We propose the Self-Explainable Affordance Learning Model\footnote{The dataset and code will be released.}, an advanced framework that seamlessly integrates self-explainable embodied caption with visual affordance learning. \\
(4) Extensive qualitative and quantitative experiments demonstrate the effectiveness and interpretability of our approach.
\section{Related Work}
\label{sec:formatting}

%-------------------------------------------------------------------------
\subsection{Visual Affordance Learning}

Visual affordance learning has emerged as an important research area in robotics, enabling robots to imitate human behavior for executing suitable actions. This task requires the model to learn the touchable regions of objects, given a large amount of human behavioural exocentric visual data. The model could predict action possibility region in objects when inputting egocentric images in reality. Previous works~\cite{zhou2016learning,mai2020erasing,chuang2018learning,pan2021unveiling,gao2021ts} focused on learning object affordance localization coordinates from image features, facilitating the robot's ability to perform basic actions like grasping and pushing, derived directly from real scenes. Differently, recent studies~\cite{liu2022joint,goyal2022human,bahl2023affordances,nagarajan2019grounded} have explored learning touchable points and trajectories from extensive human behavior videos. Some weakly supervised affordance grounding methods~\cite{luo2022learning,hadjivelichkov2023one,li2023locate} focused on leveraging large datasets of weakly labeled images-region for robust robotic grounding. Nagarajan et al.~\cite{nagarajan2019grounded} attempted to leverage videos of human behavior and associated action labels to enhance the learning of affordances in egocentric images. They employed image matching techniques to identify the behavioral class and facilitate the transfer of knowledge to generate hotspot maps for images. Luo et al.~\cite{luo2022learning}  employed affordance invariance mining to connect exocentric interactions and egocentric affordance, utilizing weak annotation and enhancing understanding of object interactions. Li et al.~\cite{li2023locate} introduced a  large image clustering to transfer knowledge from exocentric to egocentric images for enhanced affordance grounding. %Geng et al.\cite{geng2022end} proposed reinforcement learning within a 3D simulated environment to align target touch points with specific paths. Additionally, Mees et al.\cite{mees2023grounding} introduced the use of large language models (LLMs) for instruction-corresponding visual affordance learning. 

Despite these advancements, a significant gap remains in the field of embodied vision affordance, specifically in predicting touchable heat points and delivering explainable behavioral descriptions. This work aims to bridge this gap by introducing a vision-language model for self-explainable affordance learning. Our method not only enhances the prediction of touchable points, but also produces embodied self-explainable caption. 
% We propose to integrate visual predictions with forthcoming behavioral descriptions.
% The text concept learned from exocentric images can also guide the egocentric prediction.
%-------------------------------------------------------------------------
\subsection{Embodied Vision-Language}

The intersection of embodied vision and language has been actively studied in recent years, encompassing areas such as navigation~\cite{qiao2023march,hu2023explore}, object manipulation~\cite{li2023behavior,padalkar2023open}, and interactive dialogue~\cite{zhao2023chat}. The Ego4D project~\cite{grauman2022ego4d} introduced a substantial dataset featuring robot video descriptions. Nair et al.~\cite{nair2022r3m} and Yao et al.~\cite{mu2023embodiedgpt} have introduced video captions through additional camera or video inputs to narrate the robot's actions. Anwen et al.~\cite{hu2023explore} proposed integrating scene descriptions into vision-language navigation datasets, employing image captioning techniques to augment embodied navigation. 

While significant progress has been made, self-explainabled affordance learning within the field of robotic manipulation remains untouched. The new task requires to simultaneously localize the affordance regions for objects and predict the corresponding action. Distinct from extant embodied captioning~\cite{hu2023explore,zhao2023chat,mu2023embodiedgpt,butler2018embodied,tan2020towards}, our self-explainable captions require the model to describe forthcoming actions proactively from the first perspective of robot perception. This contrasts with existing methods where embodied image/video captions are dependent on an additional camera to narrate what the robot is doing and has done. This shift necessitates the robot to communicate its perceptual predictions, thus endowing it with explainability. Human can correct anticipated actions on time, greatly improving the interaction between humans and robots. This paper fills the gap by proposing the dataset and a novel model.

\section{Methodology}
\label{Embododied Affordacne Captioning}
% In this paper, we propose a novel task: Self-Explainable Affordance Learning (SEA)
In this section, we first introduce the new task: Self-Explainable Affordance Learning (SEA). Then, we delve into the details of our SEA dataset. Last, a novel SEA model to deal with this task is explained.

% In this section, we delve into the details of our Self-Explainable Affordance (SEA) dataset and models. The core aim of this initiative is to harmonize explainable textual descriptions with conventional visual affordance grounding. Such integration empowers agents to do more than just recognize touchable hotspots from visual data depicting human behavior; it also equips them to proactively articulate their planned actions.

\subsection{Self-Explainable Affordance Learning}
%\myPara{Revisit the classic setting.}
Existing methods~\cite{li2023locate,luo2022learning,chuang2018learning} require the model to learn the action possibility region in objects from the exocentric images of human-object interactions and egocentric images. The supervision available for training is visual heat maps for action region in objects.

%\myPara{The proposed setting.} 

We argue that there are various human behaviours for objects. For example, given a image of drum, both carrying drums and beating the drums are correct actions, for which the affordance regions are totally different. However, in the existing setting of affordance learning, the model only predicts a single heat map of action. Motivated by this, we innovatively introduce self-explainable language information to traditional visual affordance learning. The newly proposed task is termed as Self-Explainable Affordance learning (SEA), which requires the model to not only localise the affordance region in the object, but also predict the embodied caption associated with that object. The embodied caption refers to object-action descriptions, i.e., ``I plan to beat the drums''. 
Our task provides interpretability to existing visual affordance learning and also facilitates downstream human-computer interaction tasks.

% We argue that because of the various human behaviours in the open world, carrying drums and banging on drums may be the right behaviours in training. Therefore, we attempt to address the issue of behavioural ambiguity from a self-explanatory perspective. The model can tell the upcoming behavioural captions about interactivity regions. This provides interpretability to purely visual predictions and also builds bridges to downstream human-computer interaction feedback.  

\begin{figure*}[t]
  \centering
   \includegraphics[width=0.95\textwidth]{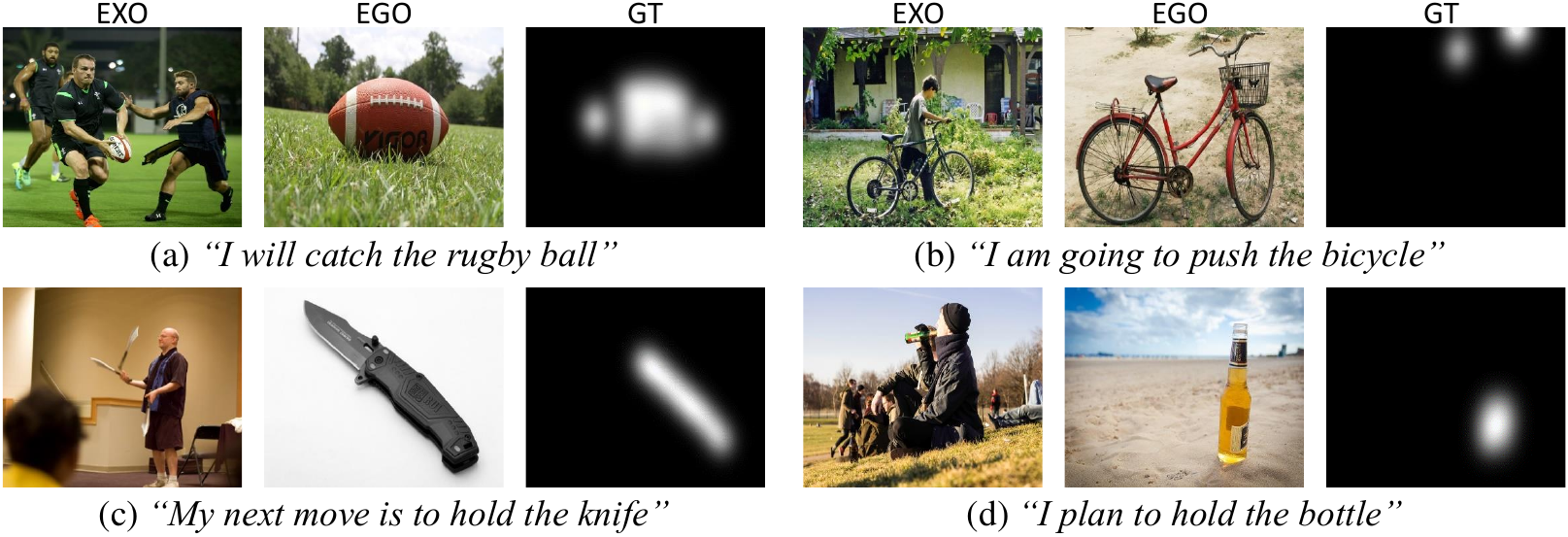}
   %\caption{In our dataset, for diverse exocentric and egocentric images, we include a range of embodied captions that articulate the agent's impending actions.}
   \caption{Examples from the SEA Dataset. Our dataset provides meaningful action captions for both exocentric and egocentric images within the realm of visual affordance learning.} %These captions are specifically designed to articulate the agent’s impending actions, exemplifying the breadth and flexibility of our self-explainable affordance learning.}
   \label{fig:example}
   \vspace{-0.5cm}
\end{figure*}

\subsection{SEA Dataset}
\label{dataset}
To facilitate the research on SEA, we develop a high-quality dataset for SEA task, aiming at training agents to simultaneously locate affordance regions in objects and predicate associated actions. It includes exocentric human-object interaction and egocentric object images, associated embodied captions, and corresponding heat maps for egocentric object images. Some examples of the SEA dataset are shown in Fig.~\ref{fig:example}.

\myPara{Image Collection.} The Affordance Grounding Dataset (AGD20K)~\cite{luo2022learning} is a large-scale dataset offering both exocentric and egocentric views, annotated with affordance maps (heat maps). It can be used for two settings~\cite{luo2022learning,li2023locate}: 1) seen setting where object categories in training and test sets are the same, and 2) unseen setting where object categories in train and test sets are different. This dataset is the main benchmark for visual affordance learning, where the model only needs to be trained to predict the heat maps of action region in objects.

For self-explainable affordance learning, the model needs to simultaneously locate the action region and predict an embodied caption. However, the existing datasets~\cite{luo2022learning} can not be used for this new task since there are not annotations of the embodied captions for objects. Therefore, we introduce the SEA dataset, which is built on AGD20K by additionally providing object-action descriptions. Initially, multiple expert teams specializing in robot vision-language research screened a vast array of visual data to identify and exclude ambiguous instances. This step included rigorous data cleaning and filtering across various classes. 

% However, diverging from these previous explainable works~\cite{chuang2018learning,zhou2017scene} that prioritize object segmentation and comprehensive image captions, our work emphasizes the generation of robotic self-explainable captions derived from heatmaps of predicted touchable points. Our dataset also can serve as a guideline for incorporating large language models in robots, especially for prompt-based learning text generation.  We manually screened, added and deleted these images from ego/exo data.

\myPara{Annotation Process.} 
% The annotation process involved two key steps. 
To generate embodied caption, the annotators looked at a pair of images including a exocentric/egocentric image and the corresponding ground-truth heat map from AGD20K, and then provide an object-action description for this sample.
% Subsequently, we performed detailed behavioral annotations, identifying different behaviors corresponding to distinct heat points on the same target. 
% In the final phase, annotators crafted expanded text descriptions based on key object-action information.
Each annotation sample was reviewed by multiple annotators to guarantee the absence of ambiguity. Annotators also performed cross-checks to verify the critical information, including location, object, and action. To ensure annotation quality, cross-annotator calibration was employed, requiring annotators to locate touchable heat points based on the provided object-action captions. Notably, our diverse annotators contributed varying linguistic styles, using different adverbs and syntax based on the object-action information, catering to an array of heterogeneous agents. As shown in Fig.~\ref{fig:example}, the exocentric/egocentric images with the same behavior are annotated from the robotic first perspective.

\begin{figure}[t]
  \centering
   \includegraphics[width=0.8\linewidth]{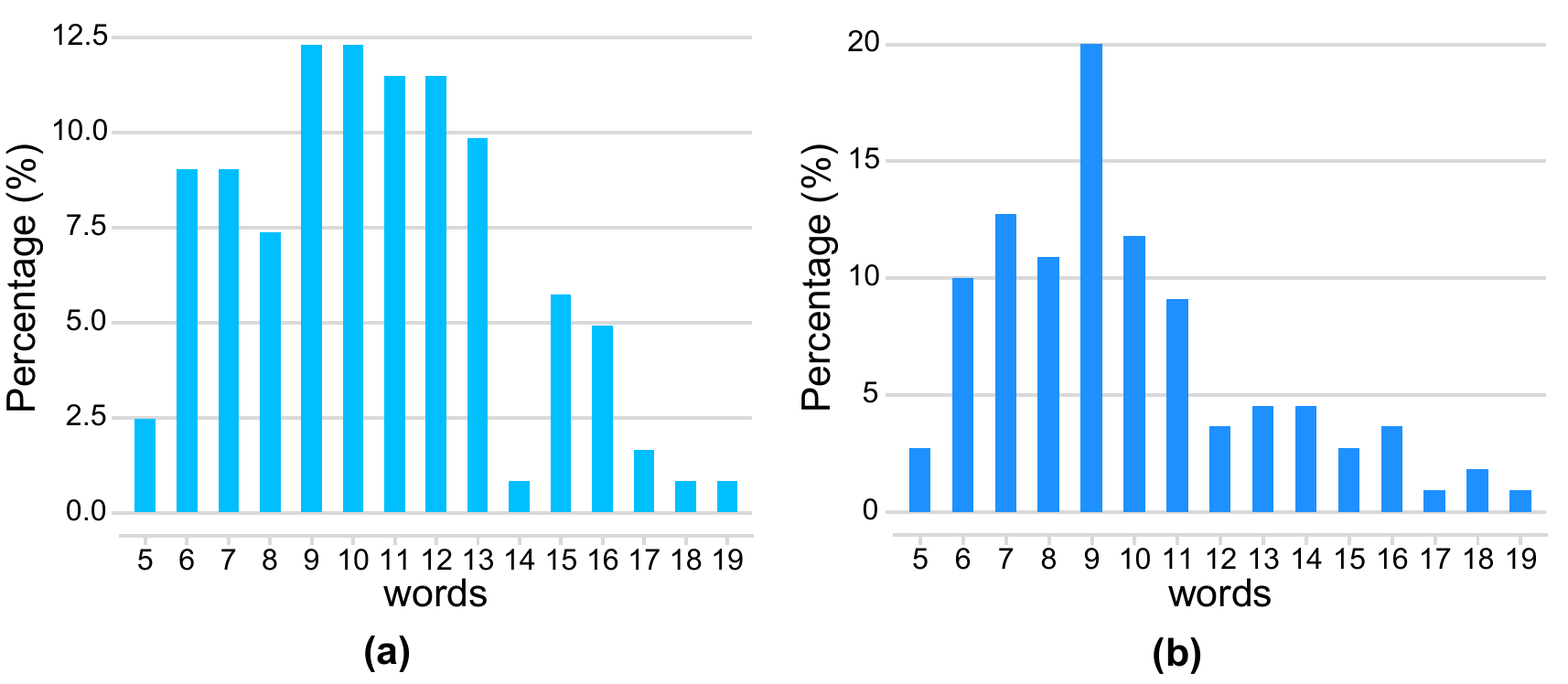}
   \caption{The statistics on the length of captions in our dataset: (a) represents the distribution of caption length in the seen scene, and (b) shows the result in the unseen scene. }
   \label{fig:data_2}
   \vspace{-0.5cm}
\end{figure}

\begin{figure}[tbp]
  \centering
  % 用resizebox来调整图像和表格的大小
  \resizebox{\textwidth}{!}{%
    \begin{minipage}{0.45\textwidth}
      \centering
      \includegraphics[width=\linewidth]{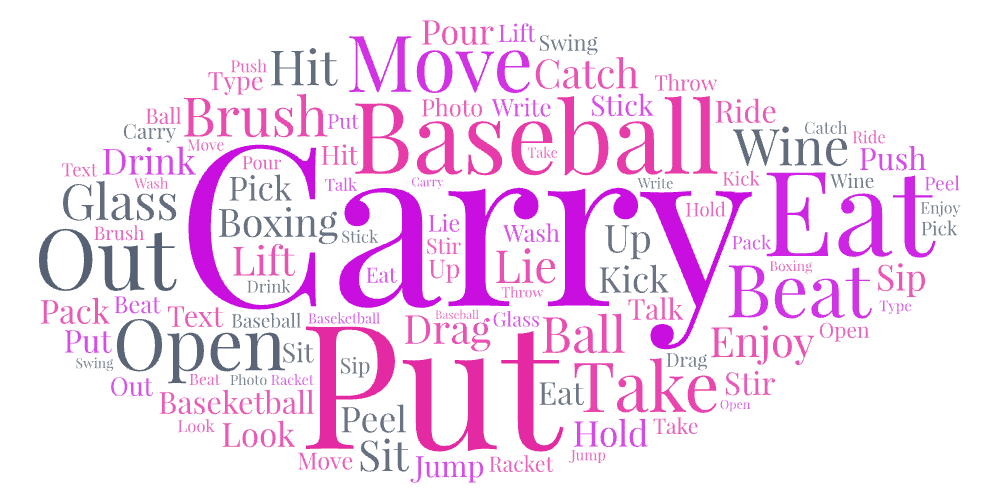}
      \vspace{-0.7cm}
      \captionof{figure}{The Word Cloud for the SEA Dataset, which displays various action and object categories in our dataset.}
      \label{fig:word}
      
    \end{minipage}
    \hfill % 添加一些水平空间
    \begin{minipage}{0.45\textwidth}
      \centering
      \captionof{table}{The statistics about the class number of our SEA dataset.}
      \vspace{0.2cm}
      \label{table:data}
      \begin{tabular}{l|c|c|c|c} % 在这里确保所有的列都被正确的竖线包围
        \hline
         &Split & Caption & Action & Object \\ \hline
        Seen & train & 3,289 & 36 & 40 \\ \cline{2-5} 
        & test & 1,573 & 36 & 40 \\ \hline
        Unseen & train & 3,756 & 25 & 34 \\ \cline{2-5} 
        & test & 1,106 & 25 & 14 \\ \hline
      \end{tabular}
    \end{minipage}%
  }
  \vspace{-0.3cm}
\end{figure}

\myPara{Dataset Details.} Fig.~\ref{fig:example} provides some examples of the SEA dataset. Table~\ref{table:data} presents the statistics of caption annotations for the SEA dataset. For the `Seen' category, we generated 3,289 training captions and 1,573 test captions. For the `Unseen' category, the dataset includes 3,756 training captions and 1,106 test captions. Furthermore, Fig.~\ref{fig:data_2} illustrates the caption word length distribution, highlighting the variation in text length in the dataset. Fig.~\ref{fig:word} shows the word cloud for objects and actions within the dataset.

Given the diversity in text styles, as evidenced in Fig.~\ref{fig:example} and Fig.~\ref{fig:word}, traditional image captioning metrics such as BLEU~\cite{papineni2002bleu} are not suitable for our dataset. Additionally, with the wide range of agents in practical applications, we have introduced an evaluation metric focusing on object-action key information classes. This metric is suitable for textual outputs from various agents, including humanoid robots~\cite{yang2022motivating,ishiguro2001robovie,kaneko2019humanoid}, five-finger robots~\cite{yamano2005five,chao2019deep,handa2020dexpilot}, the Franka robotic arm~\cite{fu2020d4rl}, among others~\cite{kornatowski2020morphing,meem2022semi}. This approach is especially relevant in the context of LLMs-based agents, which exhibit a wide range of expressions. While some methods may reduce this task to Open Vocabulary Recognition (OVR) or Open Vocabulary Classification (OVC), our goal transcends mere sentence template generation. We aim for robots to autonomously express their key behaviors in a self-explainable manner, enhancing real-world interaction and functionality.

\subsection{SEA Model}
\begin{figure*}[t]
  \centering
   \includegraphics[width=\textwidth]{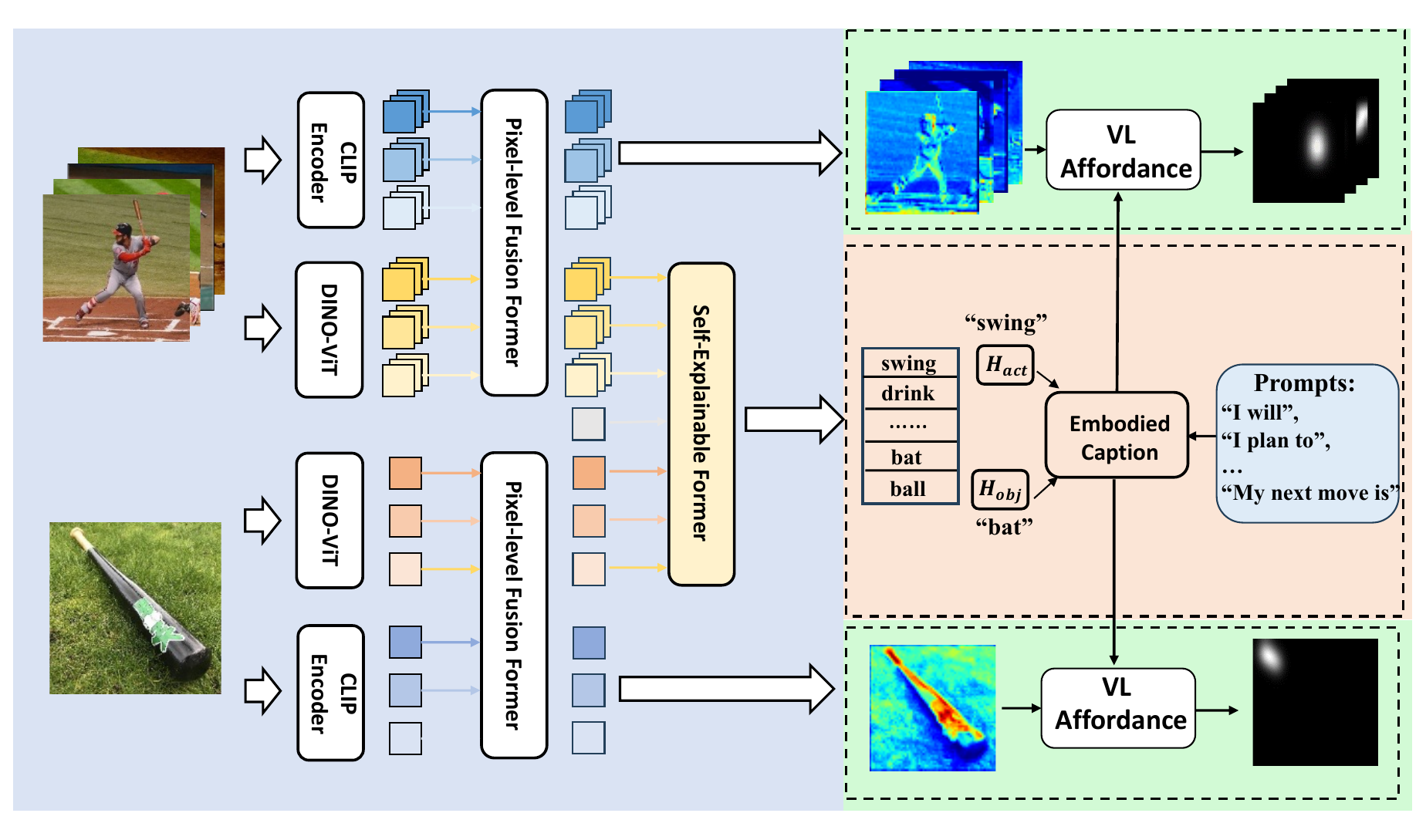}
   \caption{Overview of our proposed framework. It consists of three parts: 1. Visual Embedding from different domains 2. Self-Explainable Module and Embodied Caption. 3. VL affordance prediection. }
   \label{fig:framework}
   \vspace{-0.5cm}
\end{figure*}

\myPara{Overview.} A distinct feature of our model is its ability to self-explain the intended actions while simultaneously providing an affordance map in the visual domain. It is precisely designed to forecast the actions the agent will undertake and the objects it will manipulate, and express its behaviors in a straightforward and understandable way. 
% The primary goal of our approach is to facilitate self-explanation for the next behavior, aiding in affordance localization and resolving ambiguities commonly encountered. 
To this end, we have transformed the model’s self-explanation task into a classification problem on actions and objects, introducing the Self-Explainable Former for predicting and explaining the agent's behaviors. In addition, we incorporate a Pixel-level Fusion architecture for both the exocentric and egocentric branches. This architecture includes a pure visual encoder and a vision-language encoder, enhancing affordance map localization while maintaining classification effectiveness. Moreover, our model features novel Contrastive Learning Loss Functions, specifically tailored to foster alignment across various modalities and facilitate the transfer of knowledge. This strategy leverages behavioral information to effectively bridge the gap between different types of data, thereby enriching the model's learning capabilities. The overall framework is shown in Fig.~\ref{fig:framework}.

\myPara{Visual Embedding.} In contrast to affordance approaches~\cite{li2023locate,hadjivelichkov2023one,luo2022learning, nagarajan2019grounded},  we encode images via a pure visual backbone and a multimodal visual backbone in our framework. DINO-ViT~\cite{caron2021emerging}, a powerful self-supervised visual transformer model, is used as the pure visual backbone. CLIP~\cite{radford2021learning}, a large pre-training vision-language model, is used as the multi-modal visual backbone. Given $k$ exocentric images $I_{exo} \in \mathbb{R}^{k\times h\times w}$ and an egocentric image $I_{ego} \in \mathbb{R}^{h\times w}$, we first extract their feature map $\{F_{exo}^D, F_{ego}^D, F_{exo}^C, F_{ego}^C\}$ using CLIP and DINO-ViT respectively. These backbones are all frozen. Then, we employ a parameter-shared transformer module called Pixel-level Fusion Former (PFF) to fuse the visual features. The whole visual embedding process is as follows:
\begin{equation}\small
    \{F_{exo}^D, F_{ego}^D\} = \mathcal{E^D}(\{I_{exo}, I_{ego}\}),
\end{equation}
\begin{equation}\small
    \{F_{exo}^C, F_{ego}^C\} = \mathcal{E^C}(\{I_{exo}, I_{ego}\}),
\end{equation}
\begin{equation}\small
    \{f_{exo}^D, ..., f_{ego}^C\} = \mathcal{PFF}(\{F_{exo}^D, ..., F_{ego}^C\}),
\end{equation}
where $\mathcal{E^D}$ denotes DINO-ViT and $\mathcal{E^C}$ denotes visual encoder of CLIP, $h\times w$ means the spatial resolution of the image.

% \vspace{1pt}
\myPara{Self-Explainable Module and Embodied Caption.} After obtaining the visual features, we use the representative visual features from the DINO-ViT branch for the Self-Explaininable Module. This choice is due to DINO-ViT's extensive use of pure visual data for self-supervised learning, ensuring comprehensive coverage of pure visual information~\cite{caron2021emerging,khan2022transformers}, and its demonstrated strong performance in common recognition and classification tasks, such as ImageNet~\cite{walmer2023teaching,deng2009imagenet}. %In our framework, we also simplify the process of explaining its next actions and objects into a classification task. 

First, the features from both the exocentric and egocentric images, denoted as ${f_{exo}^D, f_{ego}^D}$, are inputted into the Self-Explainable Former. Along with these visual features, the [CLS] token is incorporated as an additional input. This token is designed to aggregate information from all image patches, playing a crucial role in subsequent classification processes. Within the Self-Explainable Former, the features from exocentric and egocentric images undergo an initial computation with corresponding source features in the self-attention layer. Subsequently, these features, along with the [CLS] token, are processed through a cross-domain attention layer. The final step involves feeding them into a Feed-Forward Network (FFN), resulting in the feature representation $f_{cls}$, which is utilized for classification tasks in the baseline model. The process is outlined as follows:

\begin{equation}\small
    \{\widetilde{f_{exo}^D}, \widetilde{f_{ego}^D}\} = \mathcal{MHSA}(\{f_{exo}^D, f_{ego}^D\}),
\end{equation}
\begin{equation}\small
    f_{cls} = \mathcal{FFN}(\mathcal{MHCA}(\{t_{cls}, \widetilde{f_{exo}^D}, \widetilde{f_{ego}^D}\})),
\end{equation}

where $\mathcal{MHSA}$ means the multi-head self-attention and $\mathcal{MHCA}$ means the multi-head cross-attention.

Subsequently, the feature representation $f_{cls}$ is directed into two separate classification heads for computation. Each head, consisting of a linear layer and a softmax function, maps $f_{cls}$ to the dimensions of action and object spaces. They then identify the corresponding action and object words with the highest scores. Furthermore, we have developed a variety of approaches for the Self-Explaining Module, utilizing techniques like Avgpool and Transformer decoder. The extracted key information is amalgamated with predefined prompts to construct complete sentences suitable for a range of agents. %For instance, as depicted in Figure~\ref{fig:example}, a generated sentence might be 'I am going to swing this baseball bat.' 
Crucially, our methodology for crafting self-explainbale captions eliminates the need for a complex, computationally demanding decoding generator. Instead, it relies on a categorization strategy that focuses on predicting essential object and action words in a straightforward way. These elements are then skillfully integrated into simple, yet informative manners such as `I will [action] [object]' or `I'm going to [action] [object],' effectively forming self-explainable text. This approach ensures the succinct and clear conveyance of vital `object-action' information. Additionally, its versatility is showcased in its adaptability to varied textual sentence structures, accommodating the diversity of robotic platforms used in real-world applications, from drones to robotic dogs.

% \vspace{1pt}
\myPara{VL Affordance Localization.} Upon generating the self-explainable outputs, our objective is to utilize the semantic information and cross-modal knowledge provided by the vision-language model to aid in localizing the affordance heatmap. Initially, we merge the identified object and action to formulate the text $t$. This text is then fed into the text encoder of CLIP, functioning as a prompt to derive the corresponding semantic representation $\widetilde{t}$. Subsequently, we calculate the cosine similarity between the visual features (both egocentric and exocentric) outputted by CLIP's visual encoder and the semantic representation $\widetilde{t}$, leading to the generation of respective heatmaps. The final step involves conducting min-max normalization on these heatmaps and applying a predefined threshold $\beta$ to refine and finalize the affordance heatmap. The process can be summarized as follows:
\begin{equation}\small
    \widetilde{t} = \mathcal{E^T}(t),
\end{equation}
\begin{equation}\small
    \{h^{exo}, h^{ego}\} = \cos(\widetilde{t}, \{f_{exo}^C, f_{ego}^C\}),
\end{equation}
\begin{equation}\small
    \{\widetilde{h^{exo}}, \widetilde{h^{ego}}\} = \mathcal{F}(\mathcal{N}(\{h^{exo}, h^{ego}\}), \beta),
\end{equation}
where $\{h^{exo}, h^{ego}\}$ are the raw affordance heatmaps. $\mathcal{N}$ means the min-max normalization. $\mathcal{F}$ means the filter computation. $\{\widetilde{h^{exo}}, \widetilde{h^{ego}}\}$ are the final affordance heatmaps.

% \vspace{1pt}
\myPara{Loss Functions.} 
In line with practices in affordance grounding~\cite{li2023locate,hadjivelichkov2023one,luo2022learning}, our model processes exocentric human-object interaction images alongside egocentric object images. The goal is to learn and transfer exocentric affordance knowledge, enabling the prediction of affordance regions in egocentric images. After obtaining the heatmaps for both exocentric and egocentric images, we also  transfer knowledge from exocentric to egocentric images in a weakly supervised manner\cite{li2023locate,luo2023grounded}.

The process begins with computing the Hadamard product of the visual feature map and the corresponding affordance heatmap. We then perform mean pooling on the weighted feature maps, aggregating them into an embedding space. To maximize the similarity between the embeddings from the two visual domains, we apply a cosine similarity loss defined as:
\begin{equation}\small
    \mathcal{L}_{cos} = max(0, 1-\alpha-\frac{ e_{exo} \cdot e_{ego}}{\left \| e_{exo} \right \| \left \| e_{ego}  \right \| }),
\end{equation}
where $e_{exo}$ and $e_{ego}$  represent the final embeddings in the exocentric and egocentric branches, respectively, while $\alpha$ is a hyper-parameter managing the domain gap.

Further, drawing inspiration from CLIP's vision-language contrastive learning, we introduce a contrastive loss between the text and visual embeddings to align and migrate external multimodal knowledge. The  matched image-text pair is positive in each training batch. This is achieved by creating negative pairs from the same batch, pushing the visual embeddings towards the direction of the matching text embedding and away from the negative text embeddings. The contrastive loss is formulated as:
\begin{equation}\small
    \mathcal{L}_i = \sum_{j=1}^{n}p_{i,j}\times \log(\frac{p_{i,j}}{q_{i,j}+\epsilon}),
\end{equation}
\begin{equation}\small
    \mathcal{L}_{con} = \frac{1}{n}\times \sum_{i=1}^{n}\mathcal{L}_i,
\end{equation}
where $p_{i,j}$ denotes the similarity between $i$-th visual embedding and $j$-th text embedding. $q_{i,j}$ denotes the ground-truth similarity. $n$ is the batch size. $\epsilon$ is a small value to avoid numerical problems. The ground truth similarity, $q_{ij}$, corresponds to the real labeled texts: $q_{ij} = 1$ for matched visual-textual features. 

\section{Experiement}
\label{sec:intro}

\begin{table*}[t]
\centering
\footnotesize
\caption{The comparison of affordance grounding ability. \ding{172} means the Self-Explain Module composed of the Feed-Forward network and Softmax function. \ding{173} consists of Concatenation and AvgPool. \ding{174} consists of Transformer.}
\resizebox{\textwidth}{!}{
\begin{tabular}{c|ccc|ccc|ccc}
\hline
 &
\multicolumn{1}{c}{\multirow{2}{*}{\begin{tabular}[c]{@{}c@{}}Text\\ Backbone\end{tabular}}} &
\multicolumn{1}{c}{\multirow{2}{*}{\begin{tabular}[c]{@{}c@{}}Self-Explainable\\ Module\end{tabular}}} &
  \multicolumn{1}{c|}{\multirow{2}{*}{\begin{tabular}[c|]{@{}c@{}}Cross-Modal\\ Loss\end{tabular}}} &
  \multicolumn{3}{c|}{Seen} &
  \multicolumn{3}{c}{Unseen} \\ 
     &           &     &  & KLD ($\downarrow$)   & SIM ($\uparrow$)   & NSS ($\uparrow$)  & KLD ($\downarrow$)  & SIM ($\uparrow$)  & NSS ($\uparrow$)  \\ \hline
Cross-view-AG~\cite{luo2022learning}&  &vision-based & & 1.524& 0.332& 0.921&  1.817& 0.272&0.817
\\ 
Cross-view-AG+~\cite{luo2023grounded}&  &vision-based & & 1.491& 0.341 &0.974 & 1.770& 0.269& 0.879 
\\ \hline
r1&BERT~\cite{devlin2018bert} &  \ding{172}   & $\times$ & 1.482 & 0.350 & 0.976 & 1.761 & 0.281 & 0.880 \\ 
r2&BERT~\cite{devlin2018bert} & \ding{173} & $\times$ & 1.418 & 0.355 & 1.009 & 1.635 & 0.343 & 1.010 \\ 
r3&CLIP~\cite{radford2021learning} & \ding{173} & $\times$ & 1.316 & 0.372 & 1.076 & 1.522 & 0.359 & 1.084 \\ 
r4&CLIP~\cite{radford2021learning} & \ding{174}  & $\times$ & 1.257 & 0.391 & 1.135 & 1.469 & 0.370 & 1.129 \\ \hline
r5&CLIP~\cite{radford2021learning} &  \ding{174}  & $\checkmark$ & \textbf{1.224} & \textbf{0.405} & \textbf{1.170} & \textbf{1.403} & \textbf{0.374} & \textbf{1.153} \\ \hline
\end{tabular}}
\vspace{-0.5cm}
\label{table:2}
\end{table*}

% \usepackage{multirow}
% \usepackage{multirow}

\subsection{Experimental Setting}

In alignment with established practices in previous work~\cite{li2023locate,mai2020erasing,pan2021unveiling,gao2021ts,nagarajan2019grounded,hadjivelichkov2023one}, our experiments were divided into two distinct settings: seen and unseen. Our model was trained with parameters set to a batch size of 16 across 20 epochs, using Stochastic Gradient Descent (SGD) for optimization. SGD with learning rate 1e-3, weight decay 5e-4. All experiments were conducted under consistent conditions, using identical hardware and parameters. The contrast loss is the prototype guidance loss, $L_{con}$. In text prediction we use cross entropy loss.

\subsection{Evaluation Metric}
\label{metric}
\textbf{Visual affordance grounding metrics.} Building upon prior affordance research~\cite{liu2022joint,fang2018demo2vec,nagarajan2019grounded}, we employ widely recognized metrics such as Kullback-Leibler Divergence (KLD), Similarity (SIM), and Normalized Scanpath Saliency (NSS) metrics, which assess the similarity between predictions and the ground truth.  In Table~\ref{table:2}, we show the experimental results of the affordance grounding evaluation.

%\myPara{Embodied self-explainable caption metrics.} In our evaluation metric, we eschew traditional text-based metrics~\cite{papineni2002bleu,lin2004rouge,vedantam2015cider} like BLEU\cite{papineni2002bleu} due to the varied nature of texts generated by various embodied agents. Instead, we prioritize the core message, assess object-action accuracy, and corroborate our analysis with top-k metrics, aligning with the nature of classification and retrieval tasks. We extract key action and target attribute information from each sentence, which could also be identified by other methods such as Named Entity Recognition (NER). '{Top-k}' accuracy means that as long as the target is within the top 'K' candidate samples, it is considered correct. 'Top-1' accuracy is equivalent to precision. The experimental results of Self-explainable descriptions are shown in table~\ref{table:3}.
\myPara{Embodied Self-Explainable Caption Metrics.} In developing our evaluation metrics, we diverge from conventional text-based approaches like BLEU~\cite{papineni2002bleu}, as motivated by the text diversity generated by various embodied agents, as detailed in Section~\ref{dataset}. Our primary focus is on capturing the core message, with a particular emphasis on object-action accuracy. This perspective is complemented by the adoption of top-k metrics, aligning well with the classification nature of our tasks. We extract vital action and target attribute information from each sentence, which can also be achieved through techniques such as Named Entity Recognition (NER)~\cite{nadeau2007survey}. In our `Top-k' accuracy metric, a target is deemed correctly identified if it ranks within the top `K' candidate samples.  The outcomes of our experiments involving Self-Explainable descriptions are detailed in Table~\ref{table:3}.

\begin{table*}[t]
\centering
\caption{The comparison of the self-explainable ability. $T_{a}@k$ means if the top-k predictions of action include ground-truth action, the prediction is right, while $T_{o}@k$ means the same on the predictions of object.}
\resizebox{0.95\textwidth}{!}{
\begin{tabular}{c|ccc|cccc|cccc}
\hline
 &
\multirow{2}{*}{\begin{tabular}[c]{@{}c@{}}Text\\ Backbone\end{tabular}} & \multirow{2}{*}{\begin{tabular}[c]{@{}c@{}}Self-Explainable\\ Module\end{tabular}} & \multirow{2}{*}{\begin{tabular}[c]{@{}c@{}}Cross-Modal\\ Loss\end{tabular}} & \multicolumn{4}{c|}{Seen} & \multicolumn{4}{c}{Unseen} \\  
& &  &  & $\text{T}_{a}@1$  & $\text{T}_{a}@5$  & $\text{T}_{o}@1$  & $\text{T}_{o}@5$ & $\text{T}_{a}@1$ & $\text{T}_{a}@5$ & $\text{T}_{o}@1$  & $\text{T}_{o}@5$  \\ \hline

r1&BERT~\cite{devlin2018bert} & \ding{172} & $\times$ & 47.4 & 70.8 & 75.6 & 86.0 & 25.6 & 48.2 & 35.1 & 53.3 \\ 
r2&BERT~\cite{devlin2018bert} & \ding{173} & $\times$ & 51.6 & 72.8 & 78.8 & 88.6 & 29.7 & 50.8 & 38.4 & 54.1 \\ 
r3&CLIP~\cite{radford2021learning} & \ding{173} & $\times$ & 57.5 & 75.3 & 80.4 & 89.2 & 36.5 & 55.1 & 48.7 & 59.3 \\ 
r4&CLIP~\cite{radford2021learning} & \ding{174} & $\times$ & 62.0 & 78.1 & 86.3 & 93.6 & 42.1 & 58.6 & 52.3 & 61.7 \\ \hline
r5&CLIP~\cite{radford2021learning} & \ding{174} & $\checkmark$ & \textbf{65.9} & \textbf{80.3} & \textbf{89.8} & \textbf{95.7} & \textbf{46.7} & \textbf{61.2} & \textbf{54.8} & \textbf{63.0} \\ \hline
\end{tabular}}
\vspace{-0.5cm}
\label{table:3}
\end{table*}

\subsection{Baselines and Comparisons}
%For our baselines, we explored two independent branches of our approach in Tables~\ref{table:2} and \ref{table:3}.
We explored multiple baseline models in Tables~\ref{table:2} and \ref{table:3}. Within the Self-Explaining module, we employed a Feed-Forward Network (FFN) and Softmax to predict the words with the highest probability. It's important to note that while the affordance heatpoint prediction module and the self-explainable caption module function as two independent branches, they share parameters in the Pixel-level Fusion Former coding fusion layer.
\begin{figure}[t]
    \centering
    \includegraphics[width=\textwidth]{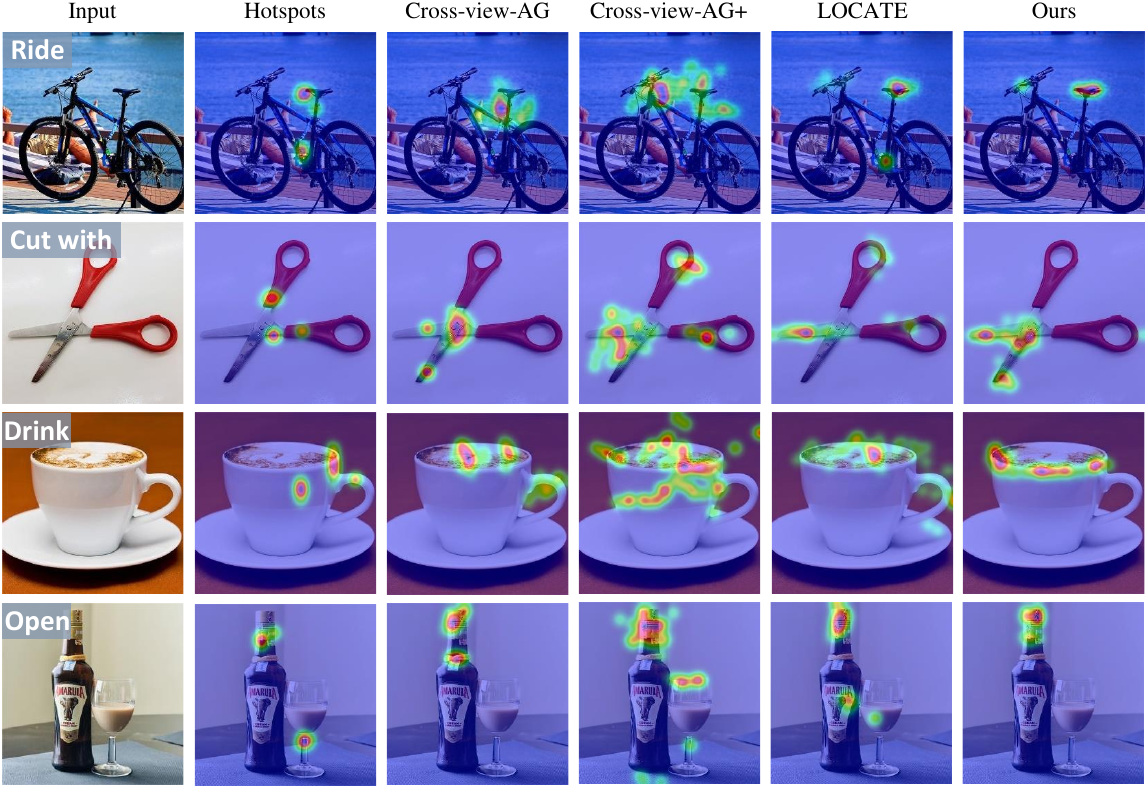}
    \caption{Visualization evaluation results. Compared to existing methods, our SEA model clearly outputs the maximum possible interaction region and significantly reduces the ambiguous output of other regions in real-world data by joint training with action-object embodied caption and affordance learning.  }
    \label{fig:enter-label}
    \vspace{-0.5cm}
\end{figure}

\begin{figure*}[t]
  \centering
   \includegraphics[width=0.98\textwidth]{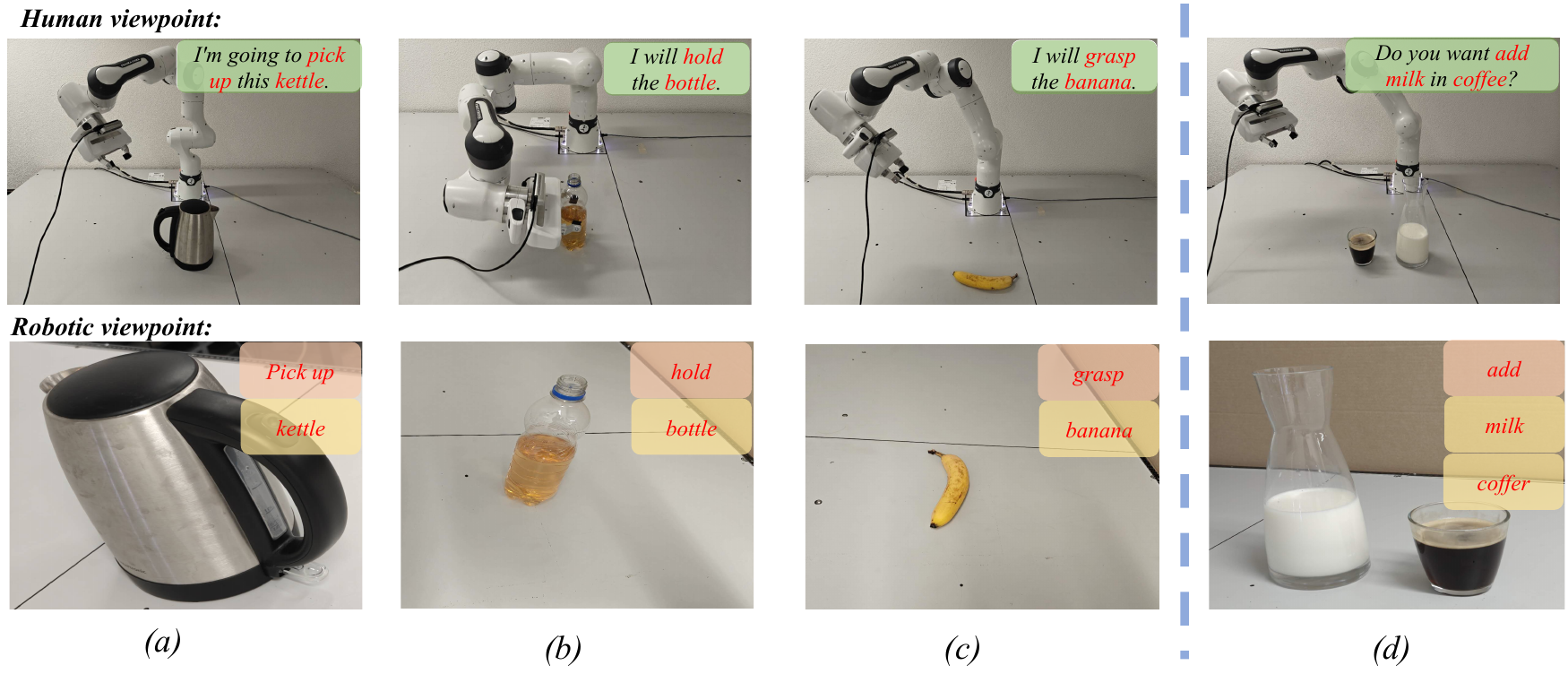}
   \caption{Visualization results of the SEA task. We intuitively demonstrate that self-explanatory embodied vision bridges human interaction with vision affordance learning.  }
   \label{fig:visual}
   \vspace{-0.5cm}
\end{figure*}

Additionally, we examined various self-explainable modules and cross-modal alignments as alternative baselines. This included the use of AvgPooling for prediction and the incorporation of the multimodal pre-training model CLIP to enhance text-image coding alignment. As depicted in Fig.~\ref{fig:framework}, a cross-modal alignment loss, the prototype guidance loss, was implemented to efficiently align the self-explainable module with the affordance heatpoint prediction module. Our SEA model demonstrated superior performance overall, as evidenced in Table~\ref{table:2} and~\ref{table:3}.

\subsection{Ablation Study}
\textbf{Affordance Grounding Ability.} The results of our experiments, r1, r2, r3, r4, including cross-modal alignment loss, self-explainable module, and backbone ablation, are detailed in Table~\ref{table:2}. %Our comparative analysis between the Bert text backbone and the CLIP backbone encoding reveals a notable improvement with the use of the VLM's pre-training backbone. 
This suggests that the VLM offers an encoding backbone enriched with external common knowledge, which is advantageous for both visual affordance grounding and self-explainable caption generation. Additionally, our findings indicate that the self-explainable module enhances robotic self-explanatory text generation, contributing to better-assisted grounding. Our SEA model, incorporating a synergistic  cross-modal alignment approach, demonstrates significant performance improvements. As shown in Fig.~\ref{fig:enter-label}, our model can reduce the ambiguity problems: multiple action regions and objects. We establish a mapping between behaviors regions and embodied behavior captions.

% \vspace{0.5cm}
\myPara{Self-Explainable Ability.} Table~\ref{table:3} illustrates the effectiveness of our approach in generating embodied self-explainable captions. As highlighted in Section~\ref{metric}, given the diversity present in heterogeneous robotic systems, we utilize the top-k evaluation metric to assess the accuracy of crucial information, specifically action and object identification. The results demonstrate that our method attains a substantial improvement under the top-k percentage measurement. Particularly in Top-1 accuracy, it reflects our method's significant capability to enhance the model's expression of key information. In Fig.~\ref{fig:enter-label},  we also show affordance comparison. SEA locates affordances based on the top-1 maximum probability behavior of visual language, thereby reducing ambiguity and increasing interpretability. This enhancement is a testament to the efficacy of cross-modal joint training. However, there remains room for improvement, particularly in handling complex scenarios like the Fig.~\ref{fig:visual} sample illustrating the weaknesses also sheds light on the existing limitations in primary robotics research when facing more complex scenarios.
\vspace{-0.5cm}
\section{Discussion and Visualization}
The visualization results presented in Fig.~\ref{fig:visual} and Fig.~\ref{fig:enter-label} offer valuable insights into the effectiveness of the SEA in real-world robotic data. Our study contributes to the field by adding interpretability to visual affordance learning, a significant advancement that could assist in promptly correcting incorrect behaviors by agents.  A key realization from our work is the complexity of real-world scenarios, often characterized as open-world environments with intricate sentence structures. This complexity underscores the need for various heterogeneous robots to not only recognize a wide range of action-object pairs but also to generate contextually appropriate sentences in diverse situations. This aspect of our research highlights the ongoing challenges and the necessity for continued advancements in the field.

\section{Conclusion}
\vspace{-0.3cm}
\label{sec:intro}
In this work, we have introduced the innovative Self-Explainable Affordance (SEA) Learning, targeting interpretability and goal ambiguity in complex environments. To support this initiative, we developed a high-quality dataset and evaluation metric tailored to this task. Our proposed model skillfully combines affordance grounding with self-explainbale captions. Extensive experiments have proven it to be effective in predicting touchable points and generating self-explainable object-action behavioral captions, addressing critical challenges in robotics research. The SEA task we introduced has wide-ranging potential applications, notably in enhancing human feedback systems and interactive modalities.

% \bibliography{main}

\appendix
\section{Self-explainabe Module}
\label{appendix:additional-experiments}
Our research aims to address ambiguities in traditional Visual Affordance Learning (VAL). By adding embodied captions to the VAL exo-ego dataset, our method can produce clear behavior descriptions to express what it will do, bridging the gap between visual affordance learning and human language. 
In traditional affordance learning, the primary objective is to learn a localization function, denoted as $\Phi$, from a substantial exo information dataset $I_{exo}$, and to predict the heat points of interaction $H_{ego}$, when presented with ego data $I_{ego}$, as illustrated in Equ.~\ref{eq1}. In contrast, our self-explainable affordance learning approach, depicted in Equ.~\ref{eq2}, aims not only to predict touchable heat points but also to elucidate which actions will be performed on which objects.

\begin{eqnarray}
H_{ego} & = & \Phi \left \{ I_{exo_{1} }, I_{exo_{2} },\cdots ,I_{exo_{n} },I_{ego} \right \},
\label{eq1}
\end{eqnarray}

\begin{eqnarray}
H_{ego}, P_{act}, P_{obj}& = & {\Phi }'  \left \{\left [ I_{exo_{1} },P_{obj_{1}},P_{act_{1}  }  \right ] ,\cdots ,\left [ I_{exo_{n} },P_{obj_{n}},P_{act_{n}  } \right ] , I_{ego} \right \} .
\label{eq2}
\end{eqnarray}

In Eq.~\ref{eq2}, ${\Phi}'$ extends the learning function to not only predict heatpoints but also identify probable actions $P_{act}$, and objects $P_{obj}$, enhancing the model's self-explanatory capability by incorporating action and object predictions along with heatpoint localizations.

\begin{figure*}[b]
    \centering
    \includegraphics[width=0.55\textwidth]{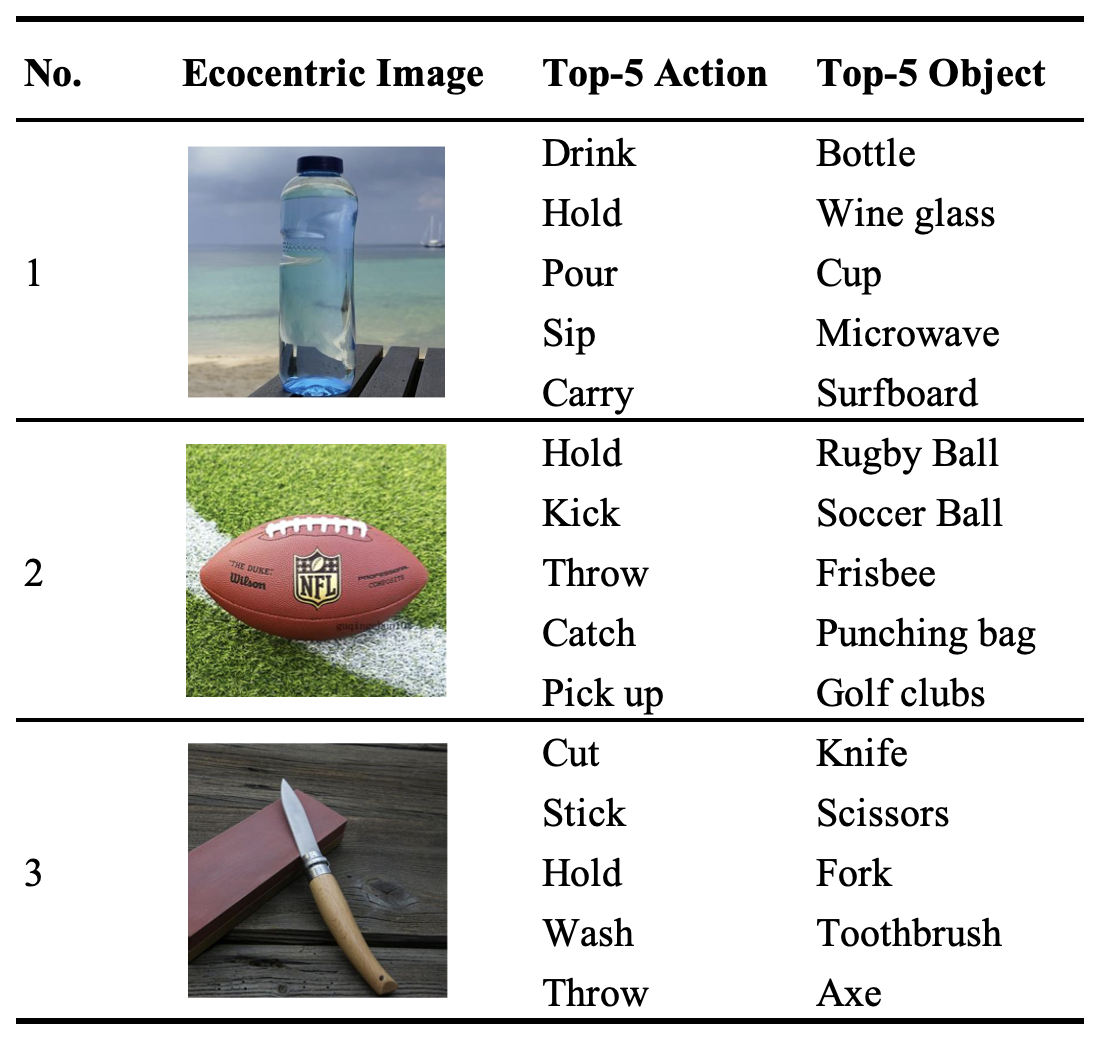}
    \caption{We present the top-K predictions for action-object combinations generated by our model. This showcases our model's ability to accurately identify and rank the most likely interactions between various actions and objects within the dataset. }
    \label{fig:top_k}
\end{figure*}

Then, we adopt straightforward but effective basic templates, such as `I will [action]+[object]', to encapsulate key information. This approach aligns with our key criteria for information judgment. We also show the results of the top k predictions in Fig.~\ref{fig:top_k}, further demonstrating its predictive proficiency. 
We use the CLIP text encoder to encode the embodied captions.

\begin{figure*}[hbt!]
    \centering
    \includegraphics[width=0.8\textwidth]{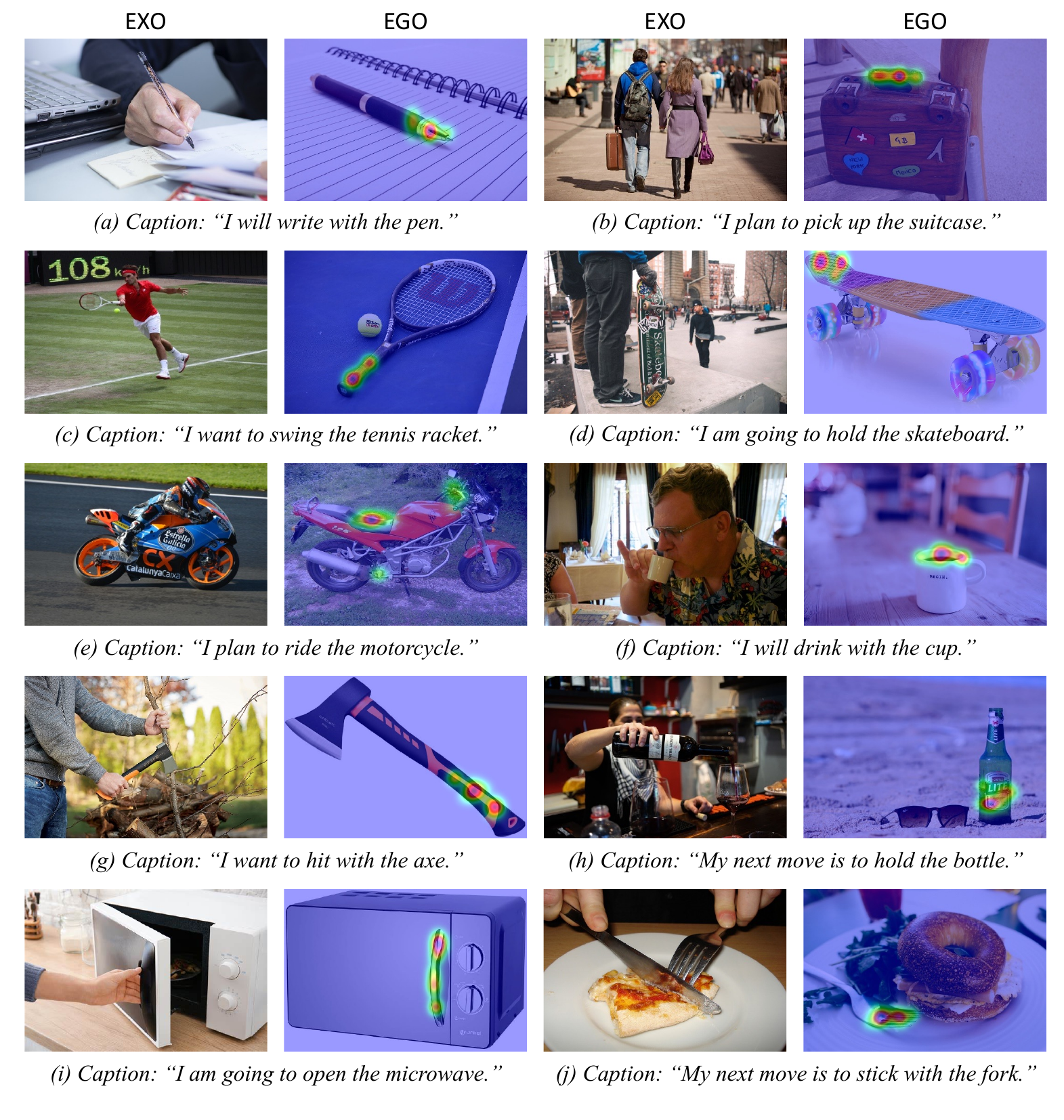}
    \caption{The self-explainable affordance learning.  }
    \label{fig2}
\end{figure*}

\section{Visualisation and Presentation}
This section is dedicated to showcasing more extensive visualizations of our SEA and the results derived from our model. 

In Fig.~\ref{fig2}, we show the self-explainable affordance leanring. Learning from human interaction exo data, it can predict touchable points on ego data and generate the corresponding self-explanatory descriptions. We also argue that self-explainable embodied captions can help visual affordances learn to locate touchable areas.

\begin{figure*}[t]
    \centering
    \includegraphics[width=\textwidth]{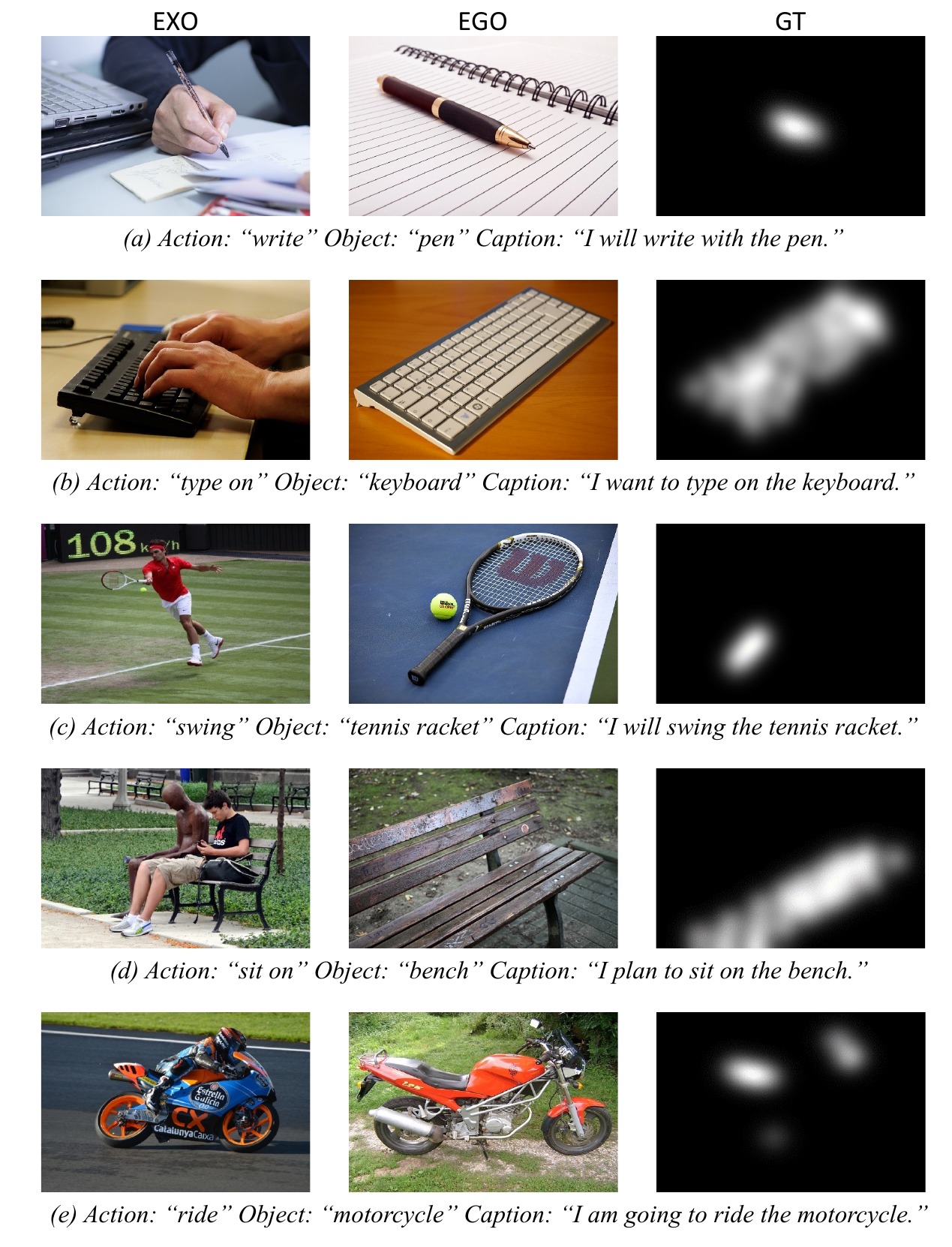}
    \caption{Visualization results of our self-explainable affordance learning. }
    \label{fig3:enter-label}
\end{figure*}

\begin{figure*}[hbt!]
    \centering
    \includegraphics[width=\textwidth]{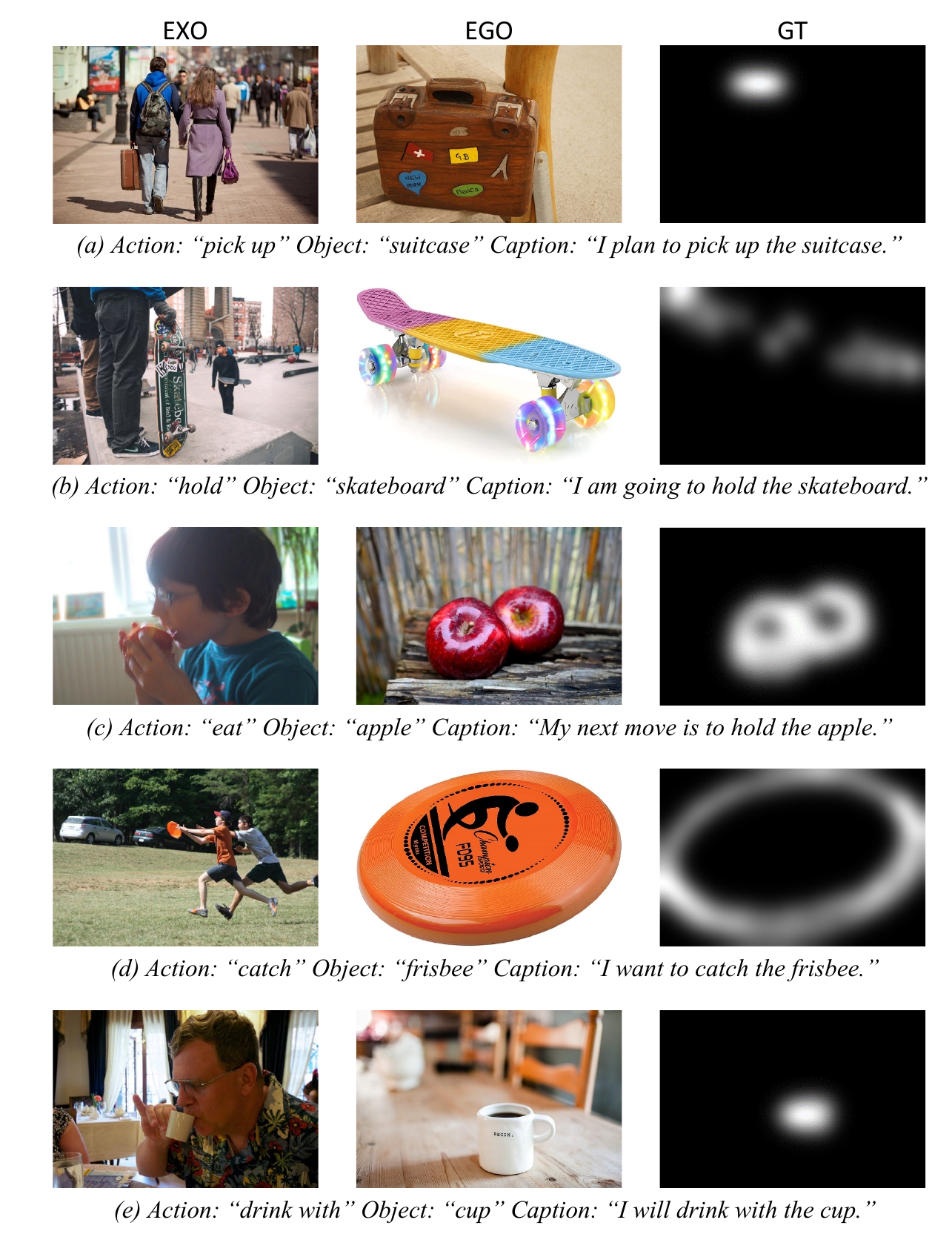}
    \caption{Visualization results of our self-explainable affordance learning.}
    \label{fig4:enter2-label}
\end{figure*}

In Fig.~\ref{fig3:enter-label} and Fig.~\ref{fig4:enter2-label}, we present visualization results that demonstrate self-explainable affordance learning. Our model is adept at deriving predictions for goal-oriented touchpoints and object-action relationships from human visual affordance data. This capability enables it to generate captions that are self-explanatory, enhancing the understanding of how it interprets and interacts with the visual data.

% \newpage

% \newpage
% \\

\end{document}